\documentclass[conference,a4]{IEEEtran}

\usepackage{times}
\usepackage{epsfig}
\usepackage{graphicx}
\usepackage{amsmath}
\usepackage{amssymb}
\usepackage{gensymb}
\usepackage{url}

\usepackage{array}
\usepackage{multirow}
\usepackage{textcomp}
\usepackage{array}
\usepackage{xcolor,colortbl}
\graphicspath{{./graphics/}}
\usepackage{subcaption}
\newcommand{\eg}{{\it e.g.},~}
\newcommand{\ie}{{\it i.e.},~}
\newcommand{\etal}{{\it et al.}~}

\begin{document}
\title{Data-Driven Segmentation \\of Post-mortem Iris Images}

\author{Mateusz Trokielewicz\\
Biometrics Laboratory\\
Research and Academic Computer Network\\
Kolska 12, 01-045 Warsaw, Poland\\
{\tt\small mateusz.trokielewicz@nask.pl}
\and
Adam Czajka\\
University of Notre Dame\\
384 Fitzpatrick Hall of Engineering\\
46556 Notre Dame, IN, USA\\
{\tt\small aczajka@nd.edu}
}
\maketitle

\begin{abstract}
This paper presents a method for segmenting iris images obtained from the deceased subjects, by training a deep convolutional neural network (DCNN) designed for the purpose of semantic segmentation. Post-mortem iris recognition has recently emerged as an alternative, or additional, method useful in forensic analysis. At the same time it poses many new challenges from the technological standpoint, one of them being the image segmentation stage, which has proven difficult to be reliably executed by conventional iris recognition methods. Our approach is based on the SegNet architecture, fine-tuned with 1,300 manually segmented post-mortem iris images taken from the Warsaw-BioBase-Post-Mortem-Iris v1.0 database. The experiments presented in this paper show that this data-driven solution is able to learn specific deformations present in post-mortem samples, which are missing from alive irises, and offers a considerable improvement over the state-of-the-art, conventional segmentation algorithm (OSIRIS): the Intersection over Union (IoU) metric was improved from 73.6\% (for OSIRIS) to 83\% (for DCNN-based presented in this paper) averaged over subject-disjoint, multiple splits of the data into train and test subsets. This paper offers the first known to us method of automatic processing of post-mortem iris images. We offer source codes with the trained DCNN that perform end-to-end segmentation of post-mortem iris images, as described in this paper. Also, we offer binary masks corresponding to manual segmentation of samples from Warsaw-BioBase-Post-Mortem-Iris v1.0 database to facilitate development of alternative methods for post-mortem iris segmentation.\let\thefootnote\relax\footnote{Manuscript accepted for publication at the IEEE International Workshop on Biometrics and Forensics (IWBF2018), \copyright 2018 IEEE}
 
\end{abstract}

\section{Introduction}
\label{sec:Introduction}

\subsection{Post-mortem iris recognition}
Post-mortem iris recognition has recently gained considerable attention in the biometric community. While this method of personal identification works nearly perfectly when applied to living individuals, it has been shown that the performance will deteriorate when existing iris recognition algorithms are confronted with images obtained in the post-mortem scenario, \ie from deceased subjects \cite{BostonPostMortem, TrokielewiczPostMortemICB2016}. This deterioration will continue as time since death elapses, due to significant distortions of the iris and the cornea caused by post-mortem decay processes, however, the first evaluations of the dynamics of post-mortem iris recognition degradation, published by Trokielewicz \etal \cite{TrokielewiczPostMortemBTAS2016}, suggests that  conventional iris recognition algorithms are able to deliver correct matches for samples acquired even 17 days after death when bodies are kept in the mortuary conditions. Bolme \etal studied the decomposition of iris, among other biometric capabilities, when the cadavers are put in an outdoor environment, simulating one of a typical forensic scenarios \cite{BolmePostMortemBTAS2016}. More recently, Sauerwein \etal \cite{Sauerwein_JFO_2017} showed in their experiments that irises stay viable up to 34 days post-mortem, when cadavers were kept in outdoor conditions and during the winter. No iris recognition method was used to support their claim, and it was based on the opinion of human experts acquiring the samples. However, it suggests that winter conditions increase the chances to see an iris even in the cadaver left outside for a longer time. All these papers suggest that automatic post-mortem iris recognition could lead to important application in forensics, being an additional tool for forensic examiners. This could help identify victims of crimes and accidents in cases when other methods of identification are unavailable or would prove more difficult to use.

\subsection{Challenges}
Erratic image segmentation is often put forward as a potential cause of degrading the performance of iris recognition algorithms when they are made to work with difficult samples, with post-mortem samples being no exception. Post-mortem decay at the cellular level slowly leads to macroscopic changes in the eye, such as deviations from the pupil's circularity, wrinkles on the cornea that cause additional specular reflections to appear, and changes in the iris texture \cite{TrokielewiczPostMortemBTAS2016}.  At the same time, it is known that the correct execution of the segmentation stage is crucial for ensuring good accuracy of iris recognition, which is conditional on encoding the actual iris texture, and not the surrounding portions of the eye. Hence, it is evident that making an iris recognition more reliable for iris image acquired after death, the segmentation methods should be designed so that to be sensitive to these new, post-mortem deformations.

\subsection{Contributions}
To our knowledge, there are no prior papers, or published research introducing the iris image processing methodologies specific to post-mortem samples. Hence, this paper is unique in the sense that it makes the first step in making post-mortem iris recognition more reliable and more attractive for forensic analyses by proposing iris segmentation specifically designed for post-mortem iris samples. This paper offers the following {\bf three contributions}: 

\begin{itemize}
	\item an algorithm for the segmentation of post-mortem iris images based on a deep convolutional neural network and experimental results showing that it offers a considerable improvement over the segmentation results produced on the same data by a conventional segmentation method,
	\item source codes of the end-to-end post-mortem segmentation method, discussed in this paper, along with the weights of the trained DCNN,
	\item manually segmented masks for Warsaw-BioBase-Post-Mortem-Iris v1.0 database \cite{WarsawColdIris1} to facilitate the development of other, post-mortem-aware segmentation methods. 
\end{itemize}

Source codes, network weights and manual segmentation results can be obtained at \verb|http://zbum.ia.pw.edu.pl/EN/node/46|. 

Structure of this study goes as follows: Section \ref{sec:Related} provides an overview of the existing applications of deep convolutional networks for the purpose of segmenting difficult iris images. Dataset, network model, its training and evaluation, and comparison against the conventional iris recognition method are described in Section \ref{sec:Experiments}. Finally, Section \ref{sec:Conclusions} discusses the main accomplishments and further work. 

\section{Related work}
\label{sec:Related}

\subsection{Deep convolutional networks for image segmentation}

Deep convolutional neural networks (DCNN) have recently shown great potential for solving selected computer vision tasks, such as natural image classification, with most popular architectures being described in \cite{AlexNet2014, VGGSimonyanCNNsForRecognition2014, ResNet2013}, and image segmentation by dense labeling, which has been reviewed extensively in \cite{SemanticSegmentationREVIEWarxiv}. These approaches are often named {\it data-driven}, as they learn the correct solution from the data itself, with minimum use of prior knowledge and with a lot of parameters (weights) and hyperparameters to be guessed directly from the samples. This opposes to {\it hand-crafted} approaches that use the prior knowledge on the subject and the training encompasses fine-tuning of not-so-many hyperparameters, when compared to data-driven algorithms. Both approaches have upsides and downsides, and data-driven models are often used when our prior knowledge on the subject is limited or difficult to be transformed into formulas possible to be applied in hand-crafted algorithms. Segmentation of post-mortem iris images is an example of such problems. One of the most successful DCNN architectures built for semantic segmentation tasks is SegNet, comprising a fully convolutional encoder-decoder architecture \cite{SegNet2016}. The encoder stage of SegNet is composed of the VGG-16 model graph. The decoder stage comprises several sets of convolution and upsampling layers, whose target is to retrieve spatial information from the encoder output, and produce a dense pixel-wise classification output of the softmax layer that is of the same size as the input image. Because of its state-of-the-art performance, including good accuracy in iris segmentation tasks \cite{JalilianCNNIrisSegmBOOK}, and recent inclusion in the MATLAB software, SegNet was chosen as a candidate network for the task described in this paper.

\subsection{Applications of convolutional networks to iris segmentation}

Regarding the applications of iris segmentation utilizing neural networks, several attempts at this task have been made, mostly aiming to improve segmentation of difficult, noisy iris images, such as these collected in visible spectrum, using low quality equipment, and pictures captured \textit{on-the-move} and \textit{at-a-distance}. 

Broussard and Ives \cite{BroussardNNsIrisSegmentation2009} employed neural networks for determining which measurements (\eg pixel value, mean, standard deviation) and which iris regions contain the most discriminatory information. This is done by training a multi-layer perceptron to identify and label an unwrapped polar iris image pixels as either belonging to the iris or not. No assumption of circularity is made, and the network serves as a multidimensional statistical classifier to combine data from multiple measurements into a binary decision for each pixel independently. Measurements for the MLP input were selected with respect to feature saliency, \ie the authors tested which ones provide the most robust features (most discriminatory power). The proposed solution is said to approach the manually annotated ground truth masks.

Liu \etal \cite{LiuICB2016CNNsForIrisSegmentation} explored hierarchical convolutional neural networks (HCNNs) and multi-scale fully convolutional neural networks (MFCNs) for the purpose of improving segmentation of noisy iris images, \eg visible light images with light reflections, blurry images captured on-the-move and/or at-a-distance, 'gaze-away' eyes, \emph{etc.}, with iris pixels being located without any \emph{a priori} knowledge or hand-crafted rules. HCNNs constructed by the authors employ hierarchical patches as input, ranging from scales small to large for capturing both local and global iris information. However, this approach is said to lack efficiency due to the sliding of the path window, which increases the computational overhead and due to the field of neurons being limited by the patch size. MFCNs, on the other hand, are reported to overcome these limitations with no sliding window (all pixel labels predicted simultaneously) and no limitation of the neuron field size. MFCNs are said to use several layers ranging from shallow to very deep, for capturing both fine and coarse details of the iris image. Experiments were performed on the UBIRIS.v2 and CASIA.v4-distance databases, comprising noisy color images acquired in unconstrained conditions and NIR at-a-distance images, respectively. MFCNs are said to use the VGG-21 model \cite{VGGSimonyanCNNsForRecognition2014}, trained for natural image classification, which is later fine-tuned using iris images with annotated masks. The following segmentation errors, defined as deviation from the ground truth segmentation by the proportion of disagreeing pixels, are obtained by the authors: 0.9\% on the UBIRIS.v2 dataset and 0.59\% on the CASIA.v4-distance dataset. Limitations include trouble with segmenting irises in images with dark skin subjects.

He \etal approached the challenge of segmenting noisy iris images obtained in the visible spectrum with a modified DeepLab CNN model which is similar to VGG-16, but with fully connected layers replaced with fully convolutional layers of kernel size 1 and an additional upscaling layer to match the output size to this of the input. The authors trained their solution on the visible spectrum iris dataset consisting of low quality samples, and reported an accuracy of 92\% IoU (Intersection over Union), which outperforms the traditional Hough transform method applied to the same data.

Similar problem is studied by Arsalan \etal \cite{ArsalanCNNVisibleIrisSegmentation2017}, where two-stage method for segmenting noisy, visible spectrum iris images is proposed, comprising of initial approximation of iris boundary with the use of classic image processing methods, and further, finer localization with a CNN composed of a modified and fine-tuned VGG-face model. The solution is shown to achieve good accuracy in segmenting irregular specular reflections.

Jalilian and Uhl employed fully convolutional encoder-decoder networks (FCEDNs) to benchmark their performance on several iris datasets, including both good and poor quality samples \cite{JalilianCNNIrisSegmBOOK}. These FCEDNs, based on the SegNet architecture, are reported to offer segmentation accuracy comparable with traditional approaches for good quality samples, and better for those of low quality. 

\subsection{Challenges in post-mortem iris image processing}

\begin{figure}[!t]
	\centering
	\includegraphics[width=0.24\textwidth]{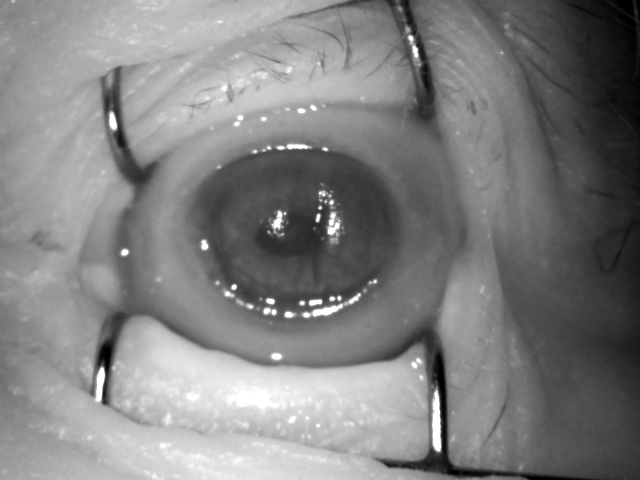}\hskip1mm
	\includegraphics[width=0.24\textwidth]{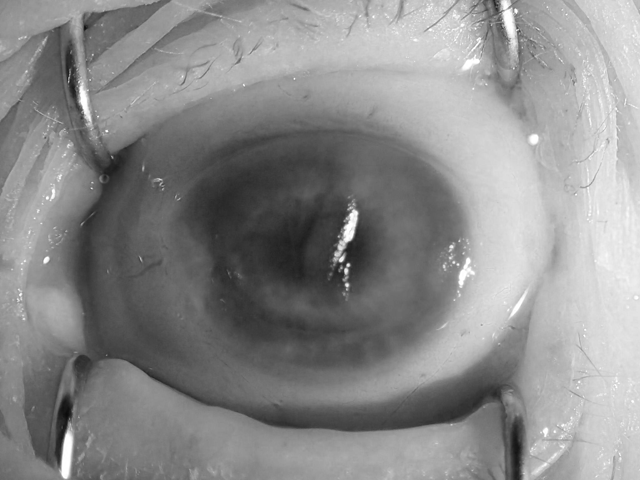}\\\vskip1mm
	\includegraphics[width=0.24\textwidth]{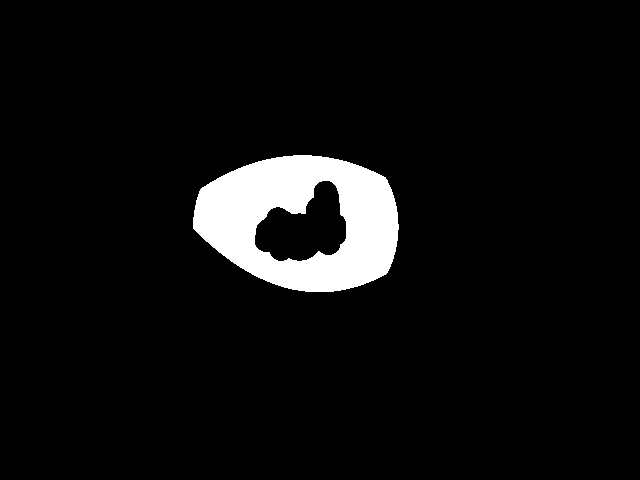}\hskip1mm
	\includegraphics[width=0.24\textwidth]{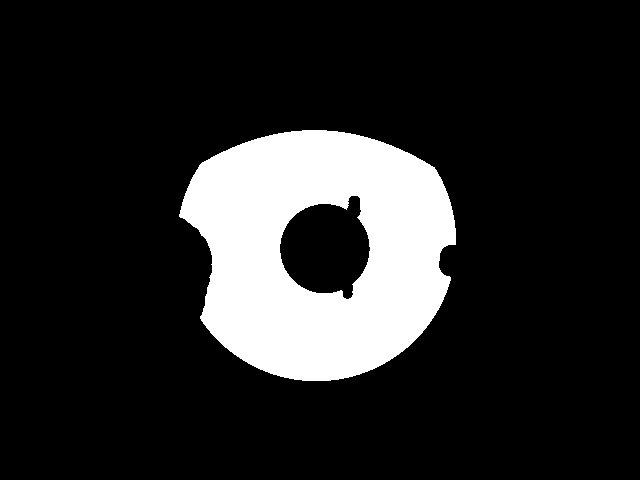}\\
	\vskip-2mm
	\caption{Example images from the Warsaw-BioBase-PostMortem-Iris-v1 dataset and their corresponding manually annotated masks, which remove the iris portions affected by post-mortem changes: NIR \textbf{(left)} and red channel of the RGB image \textbf{(right)}.}
	\label{fig:samplesFromDataset}
\end{figure}

An important conclusion that we can draw from this brief literature review, is that DCNNs built for semantic segmentation tasks are a promising solution for dealing with poor quality iris images. Post-mortem iris images represent another category of difficult iris samples since they are often heavily impacted by biological decay processes and show wrinkles on the iris texture, occurring due to excessive drying of the cornea, partial collapse of the iris due to loss of intraocular pressure, as well as additional light reflections associated with these changes. In addition to all of the above, metal retractors used to open the eyelids are often visible in the image as well, see Fig. \ref{fig:samplesFromDataset}. These make conventional iris segmentation methods, \eg those based on Daugman's idea of using circular approximations of the iris inner and outer boundaries, inaccurate and thus ineffective in algorithms targeting forensic analysis of iris samples.

\section{Experiments}
\label{sec:Experiments}

\subsection{Experimental dataset}
For the purpose of this study, we used the only, known to us, publicly available Warsaw-BioBase-PostMortem-Iris-v1 dataset, which gathers 1330 post-mortem iris images collected from 17 individuals during various times after death (from 5 hours up to 17 days) \cite{WarsawColdIris1}. These samples represent ocular regions of recently deceased subjects. Typical, near-infrared (NIR), as well as high quality visible light images are available in this dataset, and we chose to train our network using both types of samples. Careful examination of the samples shows that the nature of this data is different from any other iris dataset, with post-mortem changes being the more pronounced the more time has elapsed since a subject's demise. Apart from additional specular reflections caused by the tissue's decay, we can observe wrinkles on the cornea, haze, altered shape of the pupil, and even visible degradation of the iris tissue and partial collapses of the eyeball.  

\subsection{Preparing ground truth data}
For every sample in the dataset, we have carefully annotated the corresponding ground truth binary mask, which denotes regions of iris that are unaffected by both the post-mortem changes, as described above, and the specular reflections, regardless of their origin. Example images from the dataset and our binary ground truth masks are shown in Fig. \ref{fig:samplesFromDataset}. To expedite the training process and reduce memory overhead, the images were downsampled to the size of 120$\times$160 pixels, and the mask predictions produced by the network are later upscaled to retrieve the original size of 480$\times$640 pixels.

\subsection{Model architecture}
\begin{figure}[!htb]
	\centering
	\includegraphics[width=0.49\textwidth]{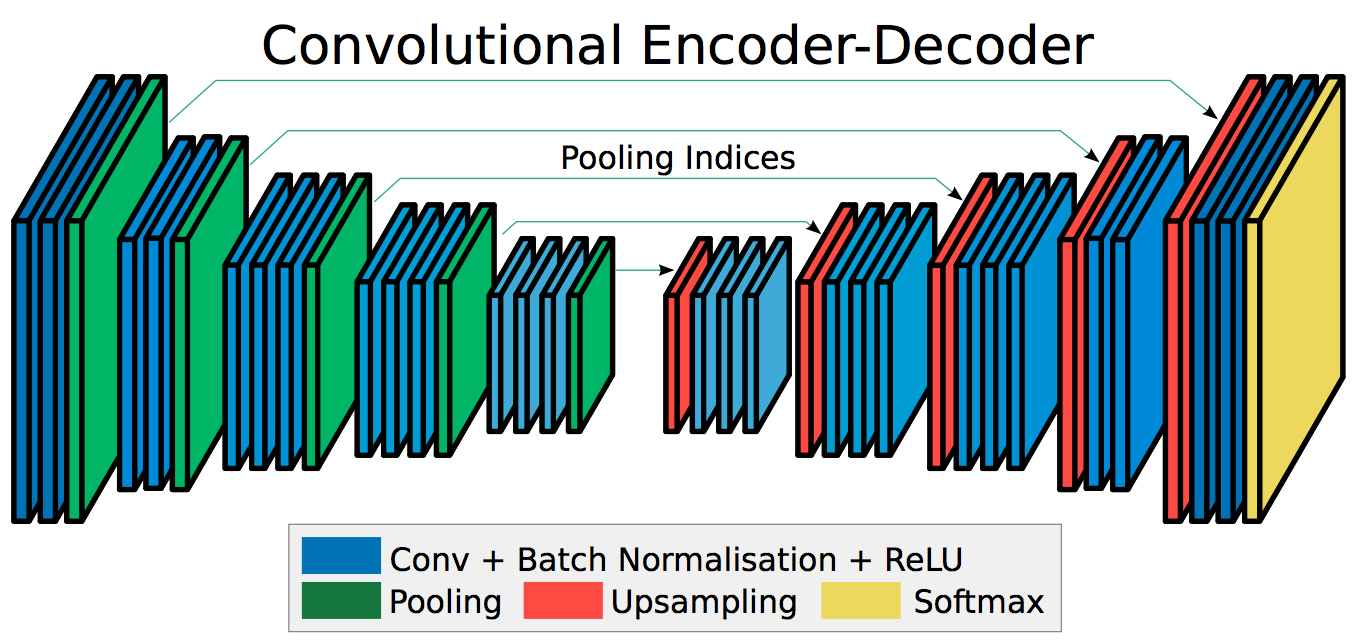}
	\caption{Encoder-decoder architecture of SegNet. Inference takes place from left to right. Size of the \emph{Softmax} layer is equal to the size of the input image. Figure adapted from \cite{SegNet2016}.}
	\label{fig:segnet}
\end{figure}

For our solution, we use the SegNet model for semantic segmentation \cite{SegNet2016}, which is a modified VGG-16 network with removed fully connected layers, and added decoder stage, so that the resulting architecture follows a concept of a coupled encoder-decoder network with five sets of convolutional and pooling/unpooling layers in each half of the network, Fig. \ref{fig:segnet}. SegNet performs the non-linear upsampling of the encoded data by employing stored indices from the max-pooling layers in a corresponding decoder. The \emph{softmax} layer is then followed by a pixel-level classification layer, which yields a binary decision for each pixel (in our case: iris or non-iris). We carried out our experiments in MATLAB 2017b environment, using the implementation of SegNet provided by the Neural Network Toolbox.     

\subsection{Training and evaluation procedure}
For training and testing procedure, 10 subject-disjoint train/test data splits were created by randomly choosing the data from 14 (out of 17) subjects to the train subset, and the data from the remaining 3 (out of 17) subjects to the test subset. All ten splits were made with replacement, making them statistically independent. The network is then trained with each train subset independently for each split, and evaluated on the corresponding test subset. This procedure gives 10 statistically independent evaluations and allows to assess the variance of the obtained results. The training, encompassing 60 epochs in each experiment, was accomplished with stochastic gradient descent as the minimization method. We applied  momentum of 0.9, learning rate of 0.001, and L2 regularization of 0.0005. During testing, a prediction in the form of binary mask is obtained from the network for each of the images. For each predicted mask, Intersection over Union (IoU) is calculated between the prediction and the ground truth mask, which is available also for test partitions of the data. These are then averaged to get the mean IoU for each test split.

\subsection{Results and comparison with conventional iris segmentation}

\begin{figure}[!htb]
	\centering
	\includegraphics[width=0.49\textwidth]{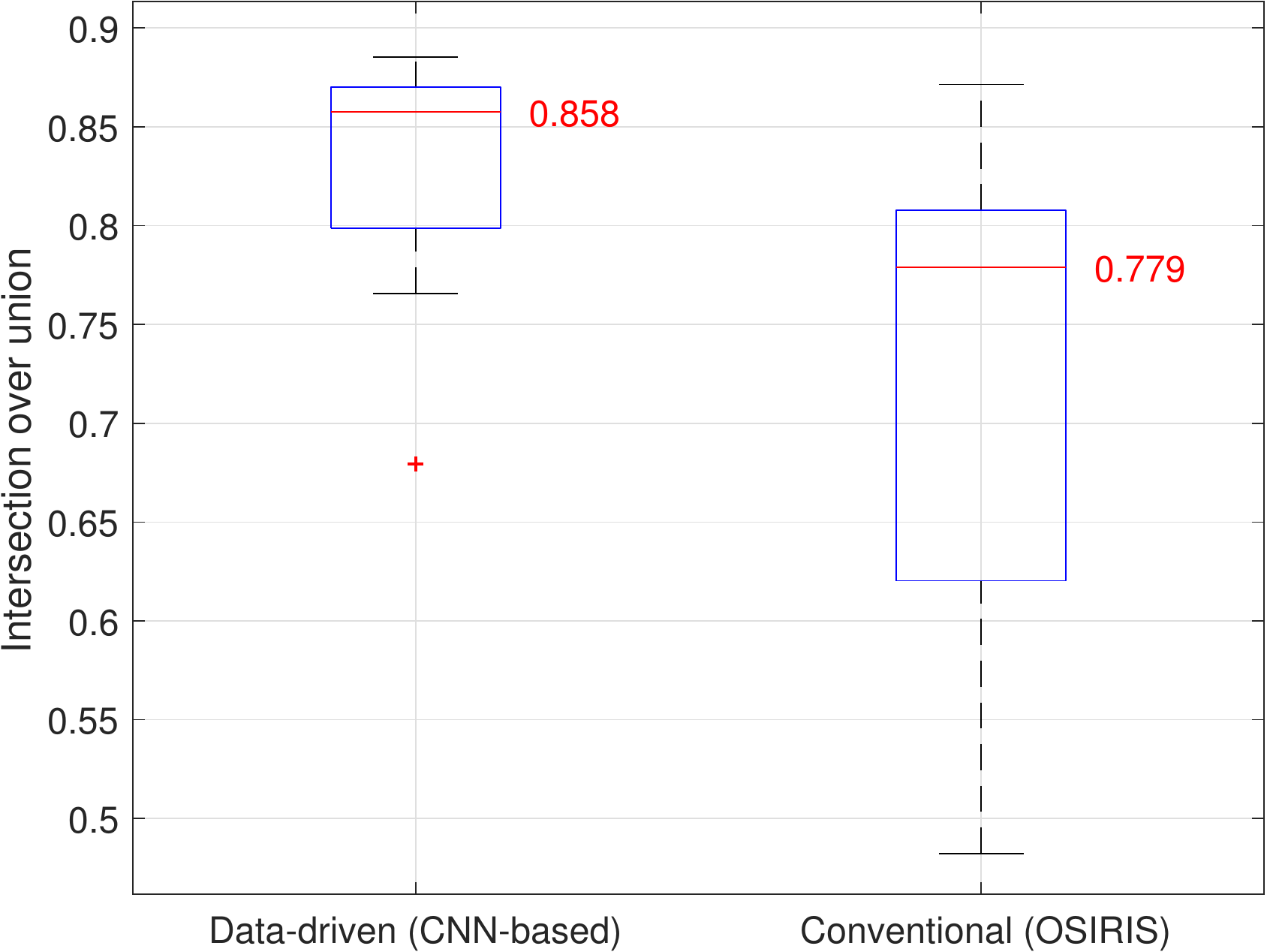}
	\caption{Boxplots representing Intersection over Union in 10 test splits, for both OSIRIS and DCNN-based approaches. Median values are shown in red, height of each boxes corresponds to an inter-quartile range (IQR) spanning from the first (Q1) to the third (Q3) quartile, whiskers span from Q1-1.5*IQR to Q3+1.5*IQR, and outliers are shown as crosses.}
	\label{fig:boxplots}
\end{figure}

To compare the DCNN-based method developed in this work with a conventional segmentation method, we did exactly the same evaluations on the train/test splits using the OSIRIS v4.1 \cite{OSIRIS} open source software that implements Daugman's idea of using circular approximations of the iris boundaries. Additionally, OSIRIS uses a Viterbi algorithm for excluding non-iris portions within the annulus defined by two circles, so it should be able to effectively cut out obstructions such as specular reflections, eyelashes and other irregular intrusions. Similarly to the evaluation of the DCNN-based solution, IoU parameters are calculated and averaged within each test split, and compared with those obtained from the DCNN-based solution. Fig. \ref {fig:boxplots} summarizes average IoU offered in all 10 splits by both solutions, and Table \ref{table:iou} details the results obtained in each split.

\begin{table}[!ht]
\renewcommand{\arraystretch}{1.4}
\caption{Mean Intersection over Union in each test split obtained for OSIRIS and DCNN-based post-mortem iris segmentation. The best and the worst results are bolded. The third column shows the improvement in performance split-wise. Averaged results are shown in the last line.}
\label{table:iou}
\centering\footnotesize
\begin{tabular}[t]{|c|c|c|c|}
\hline
& {\bf Mean IoU (OSIRIS)} & {\bf Mean IoU (CNN)} & {\bf Improvement}\\
\hline
\hline
{Split 1} & \cellcolor{red!45} 0.7793 & \cellcolor{green!45} 0.8587 & 10.2\%\\
\hline
{Split 2} & \cellcolor{red!45} 0.6002 & \cellcolor{green!45} 0.7657 & 27.6\%\\
\hline
{Split 3} & \cellcolor{red!45} 0.7786 & \cellcolor{green!45} 0.8681 & 11.5\%\\
\hline
{Split 4} & \cellcolor{red!45} 0.7533 & \cellcolor{green!45} 0.8427 & 11.9\%\\
\hline
{Split 5} & \cellcolor{red!20} {\bf 0.8715} & \cellcolor{green!100} {\bf 0.8853} & 1.6\%\\
\hline
{Split 6} & \cellcolor{red!45} 0.6203 & \cellcolor{green!45} 0.7986 & 28.7\%\\
\hline
{Split 7} & \cellcolor{red!90} {\bf 0.4823} & \cellcolor{green!20} {\bf 0.6794} & {\bf 40.9\%}\\
\hline
{Split 8} & \cellcolor{red!45} 0.8032 & \cellcolor{green!45} 0.87 & 8.3\%\\
\hline
{Split 9} & \cellcolor{red!45} 0.8078 & \cellcolor{green!45} 0.8564 & 6.0\%\\
\hline
{Split 10} & \cellcolor{red!45} 0.8621 & \cellcolor{green!45}  0.8822 & 2.3\%\\
\hline
\textbf{Average} & \cellcolor{red!45} {\bf 0.7358} & \cellcolor{green!45} {\bf 0.8303} & {\bf 12.8\%}\\
\hline
\end{tabular}
\end{table}

The DCNN-based solution proposed in this paper clearly outperforms the conventional segmentation method, not only on average for the entire experiment, but also individually in each split. It provide the segmentation accuracy as high as IoU 88.53\%, while OSIRIS offers 73.58\% IoU on average in identical evaluation. This means an average improvement of 12.8\% presented by the proposed methods over the conventional algorithm. Looking at the results obtained in each data split, the DCNN-based solution always outperforms the OSIRIS, even by as much as 40.9\% (split 7).

\begin{figure*}[t]
	\centering
	\begin{subfigure}[t]{0.32\textwidth}
		\includegraphics[width=\textwidth]{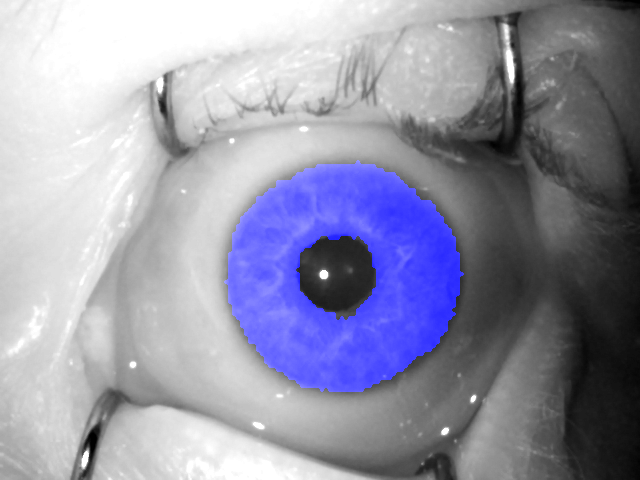}
		\caption{DCNN-based segmentation}
	\end{subfigure}  
	\begin{subfigure}[t]{0.32\textwidth}
		\includegraphics[width=\textwidth]{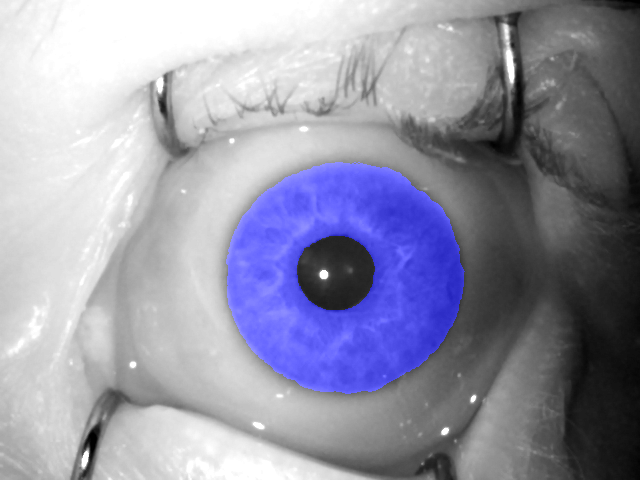}
		\caption{OSIRIS segmentation}
	\end{subfigure}  
	\begin{subfigure}[t]{0.32\textwidth}
		\includegraphics[width=\textwidth]{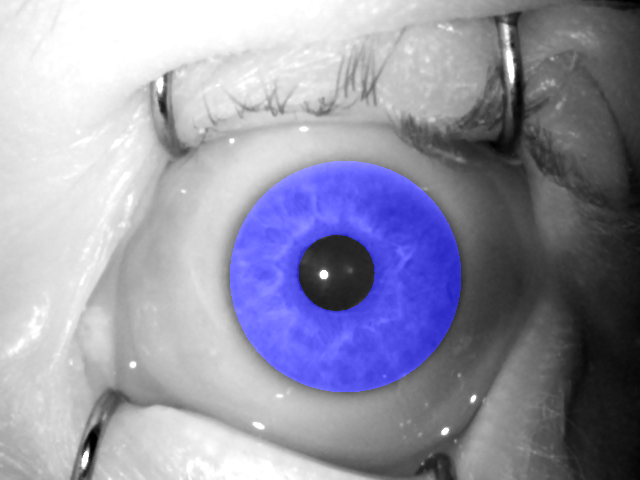}
		\caption{Ground truth}
	\end{subfigure}  
	\caption{Results of the DCNN-based (a) and the conventional (b) iris image segmentation when {\bf both methods present a good-quality outcome}. The iris image used in this example was acquired in NIR light and only 5h post-mortem. The corresponding ground truth, manually annotated on the same image is also presented (c).}
	\label{fig:example-predictions-sg-og}
\end{figure*}
% 0010_L_1_3_NIR

 \begin{figure*}[!htb]
	\centering
	\begin{subfigure}[t]{0.32\textwidth}
		\includegraphics[width=\textwidth]{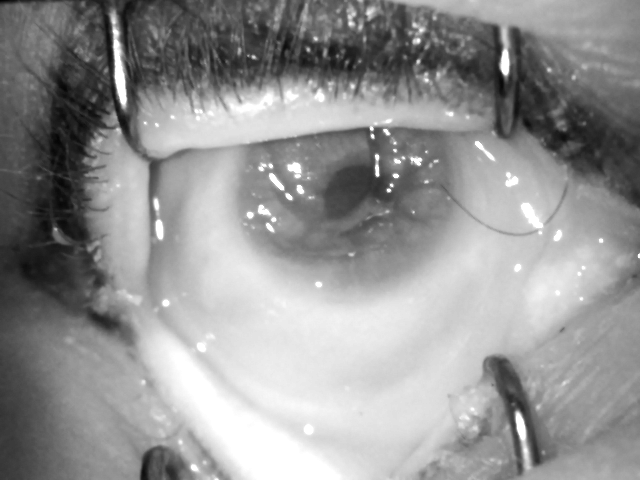}
		\caption{DCNN-based segmentation}
	\end{subfigure}  
	\begin{subfigure}[t]{0.32\textwidth}
		\includegraphics[width=\textwidth]{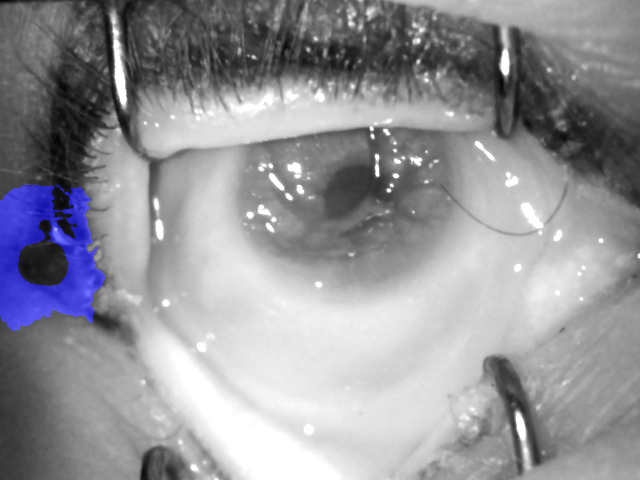}
		\caption{OSIRIS segmentation}
	\end{subfigure}  
	\begin{subfigure}[t]{0.32\textwidth}
		\includegraphics[width=\textwidth]{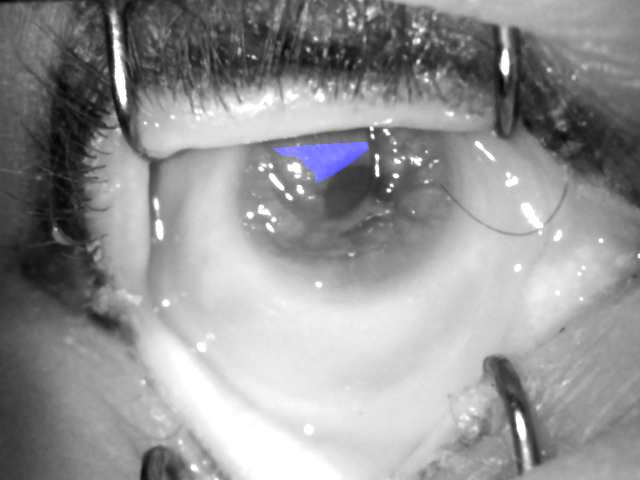}
		\caption{Ground truth}
	\end{subfigure}  
	\caption{Same as in Fig. \ref{fig:example-predictions-sg-og}, except that the failure of both the DCNN-based (a) and the conventional segmentation (b) is illustrated. The iris image used in this example was acquired in NIR light and 574 hours post-mortem.}
	\label{fig:example-predictions-sf-of}
\end{figure*}
% 0017_R_10_1_NIR 

\subsection{Close-up analysis of the results}

\begin{figure*}[!htb]
	\centering
	\begin{subfigure}[t]{0.32\textwidth}
		\includegraphics[width=\textwidth]{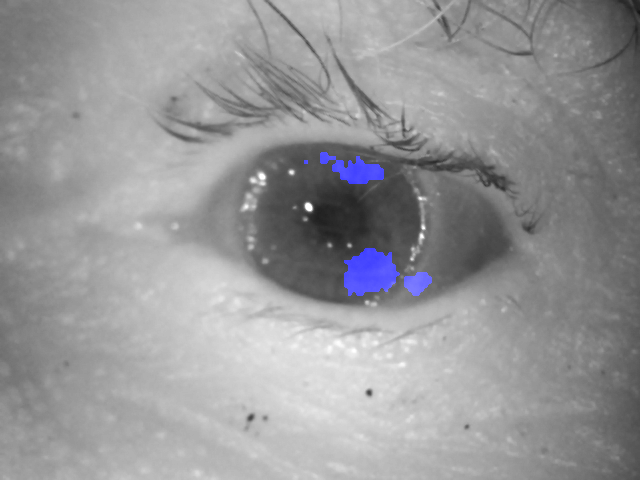}
		\caption{DCNN-based segmentation}
	\end{subfigure}  
	\begin{subfigure}[t]{0.32\textwidth}
		\includegraphics[width=\textwidth]{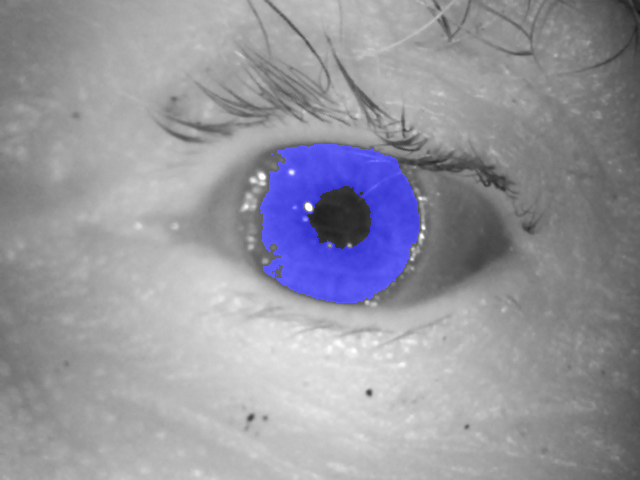}
		\caption{OSIRIS segmentation}
	\end{subfigure}  
	\begin{subfigure}[t]{0.32\textwidth}
		\includegraphics[width=\textwidth]{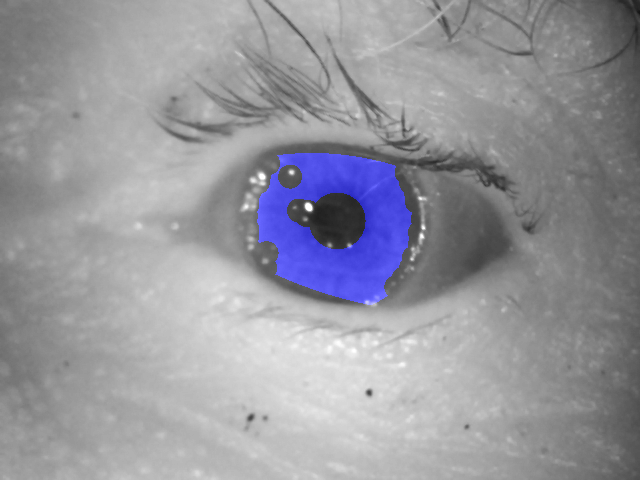}
		\caption{Ground truth}
	\end{subfigure}  
	\caption{Same as in Fig. \ref{fig:example-predictions-sg-og}, except that the failure of the DCNN-based method is presented (a), in case when the conventional segmentation did a good job (b). The iris image used in this example was acquired in NIR light and 211 hours post-mortem.}
	\label{fig:example-predictions-sf-og}
\end{figure*}

\begin{figure*}[!htb]
	\centering
	\begin{subfigure}[t]{0.32\textwidth}
		\includegraphics[width=\textwidth]{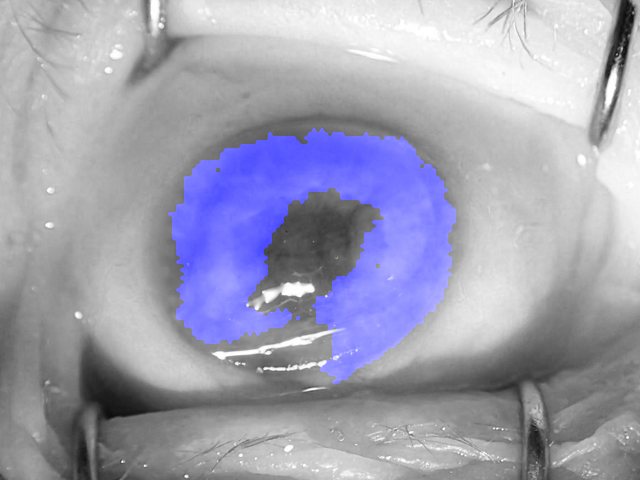}
		\caption{DCNN-based segmentation}
	\end{subfigure}  
	\begin{subfigure}[t]{0.32\textwidth}
		\includegraphics[width=\textwidth]{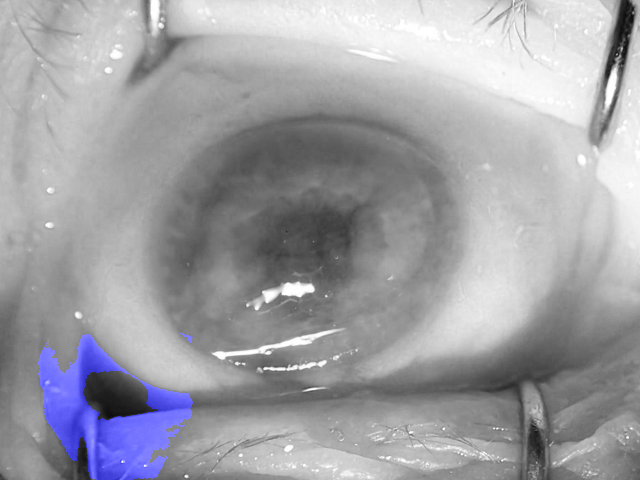}
		\caption{OSIRIS segmentation}
	\end{subfigure}  
	\begin{subfigure}[t]{0.32\textwidth}
		\includegraphics[width=\textwidth]{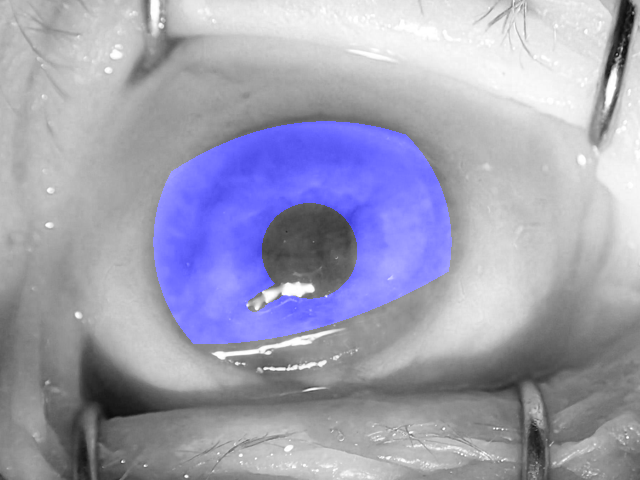}
		\caption{Ground truth}
	\end{subfigure}  
	\caption{Same as in Fig. \ref{fig:example-predictions-sg-og}, except that the results are displayed for an image presenting typical post-mortem deformations. The DCNN-based segmentation (a) did a good job when compared to the conventional segmentation (b). The iris image used in this example was acquired in NIR light and 154 hours post-mortem.}
	\label{fig:example-predictions-sg-of}
\end{figure*}

It is interesting to see example segmentation results for both DCNN-based and conventional algorithms, to discuss potential reasons of failures and room for improvement. Figures \ref{fig:example-predictions-sg-og} through \ref{fig:example-predictions-sf-of} present example segmentation results, along with ground truth annotation for comparison, in four categories: 

\begin{itemize}
	\item both algorithms performed well (achieved simultaneously the highest IoU), Fig. \ref{fig:example-predictions-sg-og},
	\item both algorithms failed (achieved simultaneously the lowest IoU), Fig. \ref{fig:example-predictions-sf-of},
	\item DCNN-based solution failed (achieved the lowest IoU) when the conventional method did a good job (achieved the highest IoU), Fig. \ref{fig:example-predictions-sf-og}, 
	\item DCNN-based solution did a good job (achieved the highest IoU) when the conventional method failed (achieved the lowest IoU), Fig. \ref{fig:example-predictions-sg-of}.
\end{itemize}

% 0018_L_4_9_NIR

% 0016_L_4_1_NIR: Pupil after death starts to deviate from its circular shape,  and the gradient between the iris and its surrounding areas is less pronounced when compared to ante-mortem samples. 

As expected, both methods perform  well for post-mortem iris images, whose quality does not diverge from a quality of alive iris images, and can be still classified as meeting the ISO/IEC 19794-6 and ISO/IEC 29794-6 requirements. Fig. \ref{fig:example-predictions-sg-og} show an example post-mortem image captured only 5 hours after death, hence in the moment when post-mortem deformations are not yet excessively present. Additionally, metal retractors used in the acquisition process made the iris texture perfectly non-occluded.

In turn, both methods failed to accurately recognize a small portion of the non-deformed iris texture in the iris that underwent heavy post-mortem processes, Fig. \ref{fig:example-predictions-sf-of}. The DCNN-based method was not able to localize any iris portion in this difficult sample acquired 574 hours (almost 24 days) post-mortem, hence producing no prediction. However, this behavior is still more favorable than what the conventional segmentation did, namely finding the iris in the incorrect region. 

There are samples which are easier to process by conventional segmentation method. Fig. \ref{fig:example-predictions-sf-og} presents a post-mortem sample that displays a regularly shaped iris with good contrast between the iris and the background. Hence, this sample was relatively easy to process by OSIRIS software, which presents a high IoU in this case. However, the intensity and texture of the iris region departed from what the DCNN saw in the training samples, and thus our solution was very selective in annotating the iris areas, ending up with low IoU.

However, one can observe an opposite result more frequently: the DCNN-based segmentation was able to detect non-standard specular reflections and wrinkles, offering way better result  than the conventional algorithm, Fig. \ref{fig:example-predictions-sg-of}. Similar results were often observed when neither the pupil nor the iris are perfectly circular, and the iris texture started to be muddy due to cornea opacification, resulting in low contrast between the iris and the surrounding areas. In such cases the supremacy of the proposed method is visible.

\section{Conclusions}
\label{sec:Conclusions}

This study presents the first known to us method for post-mortem iris image segmentation aiming at making post-mortem iris recognition more reliable. The proposed solution incorporates a deep convolutional neural network (DCNN) that already proved to be useful in semantic segmentation tasks. We presented that the DCNN-based approach is able to effectively learn deformations of the iris specific to post-mortem biological processes, and use this knowledge effectively to skip these deformed regions in the segmentation. The DCNN-based method outperforms a conventional iris segmentation algorithm by a wide margin: the Intersection over Union (IoU), averaged over 10 statistically independent experiments, equals to 83\%, where the conventional algorithm achieves IoU=73.6\%. This work thus makes the first important step in adapting iris recognition methodology to post-mortem images, opening up many new opportunities for the forensic examiners and biometrics experts.  

This paper follows the reproducibility guidelines by offering a) the source codes of the end-to-end post-mortem-aware iris segmentation method, b) trained DCNN model, and c) manual segmentation results for the publicly available post-mortem iris samples available to those who are interested in further research in post-mortem iris recognition. These, in particular, allow to fully reproduce the results presented in this paper.

\section*{Acknowledgments}
Adam Czajka acknowledges the partial support of NASK under grant agreement no. 2/2017.

The authors would like to thank Ms Ewelina Bartuzi and Ms Katarzyna Roszczewska for their help with preparing manual annotations for iris image masks. 

We are also indebted to NVIDIA for supporting us with a GPU unit that enabled this study to come to fruition.    

{\small

}

\end{document}